\documentclass{article} 
\usepackage{nips13submit_e,times}
\usepackage{url}
\usepackage{algorithm}
\usepackage{algorithmic}
\usepackage{graphicx}
\usepackage{array}
\usepackage{multirow}
\usepackage{sidecap}
\newcommand{\wrong}[1]{\textcolor{red}{#1}}
\usepackage{amsmath,amsfonts}
\usepackage{amssymb,amsopn}
\usepackage{bm} 

\newcommand{\vct}[1]{\boldsymbol{#1}} 
\newcommand{\mat}[1]{\boldsymbol{#1}} 

\newcommand{\eat}[1]{}

\usepackage[bookmarks=false]{hyperref}

\title{A Unified Semantic Embedding: \\ Relating Taxonomies and Attributes\\ }

\author{
Sung Ju Hwang\\
Disney Research \\
Pittsburgh, PA \\
\texttt{sungju.hwang@disneyresearch.com} \\
\And
Leonid Sigal \\
Disney Research \\
Pittsburgh, PA \\
\texttt{lsigal@disneyresearch.com} \\
}

\nipsfinalcopy 

\begin{document}

\maketitle

\begin{abstract}
We propose a method that learns a discriminative yet semantic space for object categorization, where we also embed auxiliary semantic entities such as supercategories and attributes. Contrary to prior work which only utilized them as side information, we explicitly embed the semantic entities into the same space where we embed categories, which enables us to represent a category as their linear combination. By exploiting such a unified model for semantics, we enforce each category to be represented by a supercategory + sparse combination of attributes, 
with an additional exclusive regularization to learn discriminative composition. 
\end{abstract}



\section{Introduction}
\vspace{-0.1in}
Semantic approaches have gained a lot of attention recently for object categorization, as object categorization problems became more focused on large-scale and fine-grained recognition tasks and datasets. Attributes~\cite{lampert-attributes,farhadi-cvpr2009,sharingfeatures,ale} and semantic taxonomies~\cite{marszalek-hierarchy,peronatree,embeddingdocument,gao11} are two of the popular semantic sources which impose certain relations between the category models. While many techniques have been introduced to utilize each of the individual semantic sources for object categorization, no unified model has been proposed to relate them.

We propose a unified semantic model where we can learn to place categories, supercategories, and attributes as points (or vectors) in a hypothetical common semantic space. Further, we propose a discriminative learning framework based on dictionary learning and large margin embedding, to learn each of these semantic entities to be well separated and pseudo-orthogonal, such that we can use them to improve visual recognition tasks such as category or attribute recognition. 

However, having semantic entities embedded into a common space is not enough to utilize the vast number of relations that exist among them. Thus, we impose a graph-based regularization between the semantic embeddings, such that each semantic embedding is regularized by sparse combination of auxiliary semantic embeddings. 

\begin{SCfigure}
\hspace{-0.1in}
\includegraphics[width=7cm]{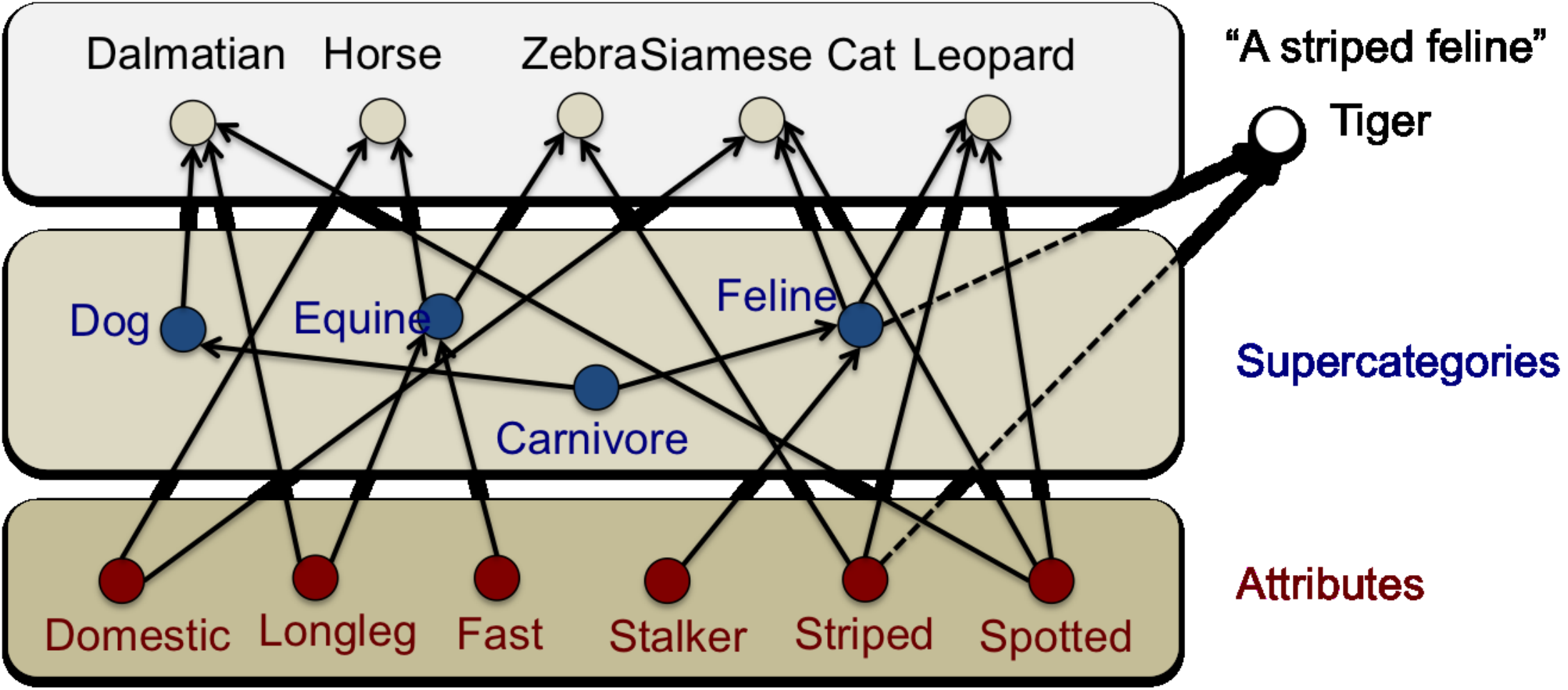}
\caption{\small \textbf{Concept:} We regularize each category to be represented by its supercategory + a sparse combination of attributes, where the combinations are learned. The resulting embedding model improves the generalization,
and is also able to compactly represent a novel category. For example, our model can describe a \emph{tiger} as a \emph{striped feline}. Such decomposition can hold for categories at any level. For example, supercategory \emph{feline} can be described as a \emph{stalking carnivore}.} 
\label{concept}
\end{SCfigure}

The observation we make to draw the relation between the categories and attributes, is that a category can be represented as the sum of its super category + the category-specific modifier, which in many cases can be represented by a combination of attributes. Further, we want the representation to be compact. Instead of describing a dalmatian as a domestic animal with a lean body, four legs, a long tail, and spots, it is more efficient to say it is a spotted dog (Figure~\ref{concept}). It is also more exact since the higher-level category dog contains all general properties of different dog breeds, including indescribable dog-specific properties, such as the shape of the head, and its posture. This exemplifies how a human would describe an object, to efficiently communicate and understand the concept.

This additional requirement imposed on the discriminative learning model would guide the learning such that we obtain not just the optimal model for class discrimination, but to learn a semantically plausible model which has a potential to be more robust and human-interpretable; we call this model Unified Semantic Embedding (USE).

\section{Learning a unified semantic embedding space}
Suppose we have $d$-dimensional image descriptors and $m$-dimensional label vectors, including category labels, at different semantic granularities, and attributes. Our goal is to embed both images and labels into a single unified semantic space.
%
To formally state the problem, given a training set $\mathcal{D}$ that has $N$ labeled examples, i.e. $\mathcal{D} = \{\vct{x}_i, y_i\}_{i=1}^N$, where $\vct{x}_i\in\mathbb{R}^d$ denotes image descriptors and $y_i\in\{1,\dots,m\}$ are their labels associated with $m$ unique concepts, we want to embed each $\vct{x}_i$ as $\vct{z}_i$, and each label $y_i$ as $\vct{u}_{y_i}$ in the $d_e$-dimensional space, such that the similarity between $\vct{z}_i$ and $\vct{u}_{y_i}$, $S(\vct{z}_i,\vct{u}_{y_i})$, is maximized. 
Assuming linear embedding with matrix $\mat{W}$, $\vct{z}_i = \mat{W}\vct{x}_i$.

\eat{
One way to solve the above problem is to use regression, using $S(\vct{z},\vct{u}_{y_i}) = \|\vct{z}-\vct{u}_{y_i}\|$. That is, we estimate the data embedding $\vct{z}_i$ as $\vct{z}_i = \mat{W}\vct{x}_i$, and minimize their distances to the correct label embeddings $\vct{u}_{y_i}$, defined as a unit vector $\vct{e}_{y_i}\in\mathbb{R}^m$ where the dimension for $y_i$ is set to 1 and every other dimension is set to 0: 
\vspace{-0.1in}
\begin{equation}
\underset{\mat{W}}{\text{min}}\sum_{c=1}^m\sum_{i=1}^N\|\mat{W}\vct{x}_i-\vct{u}_{y_i}\|_2^2 + \lambda \|\mat{W}\|_F^2 
\label{ridge}
\vspace{-0.05in}
\end{equation}
The above ridge regression will project each instance close to its correct embedding. However, it does not guarantee that the resulting embeddings are well separated. 
}

To ensure that the projected instances have higher similarity to its own category embedding than to others, we add discriminate constraints, which are large-margin constraints on distance:
$\|\mat{W}\vct{x}_i - \vct{u}_{y_i}\|_2^2 + 1 \le \|\mat{W}\vct{x}_i - \vct{u}_c\|_2^2 + \xi_{ic}, y_i \ne c$.
This translates to the following discriminative loss:
\begin{equation}
\mathcal{L}_C(\mat{W},\mat{U},\vct{x}_i,y_i) =  \sum_c[1 + \|\mat{W}\vct{x}_i - \vct{u}_{y_i}\|_2^2 - \|\mat{W}\vct{x}_i - \vct{u}_c\|_2^2]_+, \forall c\ne y_i,
\vspace{-0.1in}
\end{equation}
where $\mat{U}$ is the columwise concatenation label embedding vectors, such that $\vct{u}_j$ denotes $j_{th}$ column of $\mat{U}$. After replacing the generative loss in the ridge regression formula with the discriminative loss, we get the following discriminative learning problem:
\vspace{-0.1in}
\begin{equation}
\begin{aligned}
\underset{\mat{W},\mat{U}}{\text{min}}\sum_i^N\mathcal{L}_C(\mat{W},\mat{U},\vct{x}_i,y_i)+ \lambda \|\mat{W}\|_F^2 + \lambda \|\mat{U}\|_F^2, y_i \in \{1,\dots,m\},\\
\end{aligned}
\label{lme}
\end{equation}
where $\lambda$ regularizes $\mat{W}$ and $\mat{U}$ from going to infinity. This is one of the most common objectives used for learning discriminative category embeddings for multi-class classification~\cite{embeddingtree,embeddingdocument}, while ranking loss-based~\cite{wsabie} models have been also explored for $\mathcal{L}_C$. 


\vspace{-0.1in}
\paragraph{Supercategories.}
While our objective is to better categorize \emph{entry} level categories, categories in general can appear in different semantic granularities. For example, a \emph{zebra} could be both an \emph{equus}, and an \emph{odd-toed ungulate}. To learn the embeddings for the supercategories, we map each data instance to be closer to its correct supercategory embedding than to its siblings: $\|\mat{W}\vct{x}_i - \vct{u}_{s}\|_2^2 + 1 \le \|\mat{W}\vct{x}_i - \vct{u}_c\|_2^2 + \xi_{sc}, \forall{s \in \mathcal{P}_{y_i}} \text{ and } c \in \mathcal{S}_s$
where $\mathcal{P}_{y_i}$ denotes the set of superclasses at all levels for class $s$, and $\mathcal{S}_s$ is the set of its siblings. The constraints can be translated into the following loss:
\begin{equation}
\mathcal{L}_S(\mat{W},\mat{U},\vct{x}_i,y_i) =  \sum_{s\in\mathcal{P}_{y_i}}\sum_{c\in\mathcal{S}_s}[1 + \|\mat{W}\vct{x}_i - \vct{u}_{s}\|_2^2 - \|\mat{W}\vct{x}_i - \vct{u}_c\|_2^2]_+.
\vspace{-0.1in}
\end{equation}

\paragraph{Attributes.}
Attributes can be considered as a normalized basis vectors for the semantic space, whose combination represents a category. 
Basically, we want to maximize the correlation between the projected instance that possess the attribute, and its correct attribute embedding, as follows:

\begin{equation}
\mathcal{L}_A(\mat{W},\mat{U},\vct{x}_i,y_i) = \sum_{a\in \mathcal{A}_{y_i}}[\sigma-(\mat{W}\vct{x}_i)^\mathsf{T}y_i^a\vct{u}_a]_+, \|\vct{u}_a\|^2 \le 1, y_i^a \in \{0,1\},
\end{equation}
where $\mathcal{A}_c$ is the set of all attributes for class $c$, $\sigma$ is the margin (we simply use a fixed value of $\sigma = 1$), $y_i^a$ is the label indicating presence/absence of each attribute $a$ for the $i_{th}$ training instance, and $\vct{u}_a$ is the embedding vector for attribute $a$.

\paragraph{Semantic regularization.}
The previous multi-task formulation enables to implicitly associate the semantic entities, with the shared data embedding $\mat{W}$. However, we want to further explicitly impose structural regularization on the semantic embeddings $\mat{U}$, based on the intuition that an object class can be represented as its parent level class + a sparse combination of attribute as follows:
\eat{
\begin{equation}
\vct{u}_c = \vct{u}_p + \mat{U}^A\vct{\beta}_c, c\in\mathcal{C}_p, \|\vct{\beta}_c\|_0 \preceq \gamma_1, \vct{\beta}_c \succeq 0, \forall{c} \in \{1,\dots,\mathsf{C}+\mathsf{S}\}
\label{comb}
\vspace{-0.1in}
\end{equation}

where $\mat{U}^{A}$ is the aggregation of all attribute embeddings $\{\vct{u}_a\}$, $\mathcal{C}_p$ is the set of children classes for class $p$, $\gamma_1$ is the sparsity parameter, and $\mathsf{C}$ is the number of categories. We require $\vct{\beta}$ to be non-negative, since it makes more sense and more efficient to describe an object with attributes that it has, rather than it does not have.
}

\vspace{-0.2in}
\begin{equation}
\begin{aligned}
\mathcal{R}(\mat{U},\mat{B}) = \sum_c^\mathsf{C} \|\vct{u}_c - \vct{u}_p - \mat{U}^A\vct{\beta}_c\|_2^2 + \gamma_2 \|\vct{\beta}_{c} + \vct{\beta}_{o}\|_2^2;\\
c \in \mathcal{C}_p, o\in\mathcal{P}_c \cup \mathcal{S}_c, 0\preceq \vct{\beta}_c \preceq \gamma_1, \forall{c,p} \in \{1,\dots,\mathsf{C}+\mathsf{S}\},\\
\label{comb2}
\end{aligned}
\vspace{-0.3in}
\end{equation}

where $\mat{U}^{A}$ is the aggregation of all attribute embeddings $\{\vct{u}_a\}$, $\mathcal{C}_p$ is the set of children classes for class $p$, $\gamma_1$ is the sparsity parameter, and $\mathsf{C}$ is the number of categories.  $\mat{B}$ is the matrix whose $j_{th}$ column vector $\vct{\beta}_j$ is the reconstruction weight for class $j$, $\mathcal{S}_c$ is the set of all sibling classes for class $c$, and $\gamma_2$ is the parameters to enforce exclusivity. We require $\vct{\beta}$ to be non-negative, since it makes more sense to describe an object with attributes that it \emph{has}, rather than attributes it does \emph{not} have.

The exclusive regularization term is used to prevent the semantic reconstruction $\vct{\beta}_c$ for class $c$ from fitting to the same attributes fitted by its parents and siblings. Such regularization will enforce the categories to be `semantically' discriminated as well. 
With the sparsity regularization enforced by $\gamma_1$, the simple sum of the two weights will prevent the two (super)categories from having high weight for a single attribute, which will let each category embedding to fit to exclusive attributes. 

\vspace{-0.1in}
\paragraph{Unified semantic embeddings with semantic regularization.}
After augmenting the categorization objective in Eq.~\ref{lme} with the superclass and attributes loss and the sparse-coding based regularization in Eq.~\ref{comb2}, we obtain the following multitask learning formulation:
\vspace{-0.05in}
\begin{equation}
\small
\begin{aligned}
\underset{\mat{W},\mat{U}, \mat{B}}{\text{min}} \sum_{i=1}^N {\mathcal{L}_C(\mat{W},\mat{U},\vct{x}_i,y_i) + \mu_1\left(\mathcal{L}_S(\mat{W},\mat{U},\vct{x}_i,y_i) + \mathcal{L}_A(\mat{W},\mat{U},\vct{x}_i,y_i)\right)} + \mu_2 \mathcal{R}(\mat{U},\mat{B});\\
\|\vct{w}_j\|_2^2 \le \lambda, \|\vct{u}_k\|_2^2\le \lambda, 0\preceq \vct{\beta}_c \preceq \gamma_1 \forall{j}\in\{1,\dots,d\}, \forall{k}\in\{1,\dots,m\}, \forall{c,p}\in\{1,\dots,\mathsf{C}+\mathsf{S}\},\\
\end{aligned}
\label{use}
\end{equation}
where $\mathsf{S}$ is the number of supercategories, $\vct{w}_j$ is $\mat{W}$'s $j_{th}$ column, and $\mu_1$ and $\mu_2$ are parameters to balance between the main and auxiliary tasks, and discriminative and generative objective.

Eq.~\ref{use} can also be used for knowledge transfer when learning a model for a novel set of categories, by replacing $\mat{U}^{A}$ in $\mathcal{R}(\mat{U},\mat{B})$ with $\mat{U}^{\mathcal{S}}$, learned on class set $\mathcal{S}$ to transfer the knowledge from.

\eat{
\subsection{Knowledge transfer for novel categories}
We can also use the learned superclass and attributes models as a means to transfer knowledge when learning a categorization model for new categories. By enforcing the new categories to be also generated as a combination of a superclass + attributes, we can learn a more robust model especially when the training instances are scarce for the new dataset.

\begin{equation}
\small
\begin{aligned}
\underset{\mat{W},\mat{U}, \mat{B}}{\text{min}}\mu\sum_{i=1}^N \mathcal{L}(\mat{W},\mat{U},\vct{x}_i, \vct{y}_i) + (1-\mu) \mathcal{G}(\mat{U}^{prior},\mat{B})\\
\|\mat{W}_i^2 \le \lambda, \|\mat{U}_i\le \lambda\\
\end{aligned}
\label{use}
\end{equation}
}

\paragraph {Numerical optimization.} Eq.~\ref{use} is not jointly convex, and has both discriminative and generative terms. The problem is similar to the problem in~\cite{supervised-dl}, and can be optimized using a similar alternating optimization, while alternating between the following two convex sub-problems: 1) Optimization of the data embedding $\mat{W}$ and parameters $\mat{B}$, and 2) Optimization of the category embedding $\mat{U}$. 

\vspace{-0.1in}
\section{Results}
We validate our method for multiclass categorization performance and knowledge transfer on the Animals with Attributes dataset~\cite{lampert-attributes}, which consists of $30,475$ images on $50$ animal classes, with $85$ class-level attributes~\footnote{Attributes are defined on color (black, orange), texture (stripes, spots), parts (longneck, hooves), and other high-level behavioral properties (slow, hibernate, domestic) of the animals.}. We use the Wordnet hierarchy to generate supercategories. Since there is no fixed training/test split, we use \{30,30,30\} random split for training/validation/test.  For the features, we use the provided $4096$-D DeCAF features obtained from a deep convolutional neural network.


We compare USE against multiple existing embedding-based categorization approaches, that either do not use any semantic information, or use semantic information but do not explicitly embed semantic entities. For non-semantic baselines, we use 
\textbf{Ridge Regression,} a linear regression with $\ell$-2 norm, 
and \textbf{LME,} a base large-margin embedding (Eq.~\ref{lme}) solved using alternating optimization. 
For implicit semantic baselines, we consider \textbf{ALE, HLE}, and \textbf{AHLE}, which are our implementation of Akata et al.~\cite{ale}. The method inputs the semantic information by representing each class with structured labels that indicate the class' association with superclasses and attributes. We implement variants that use attributes (ALE), leaf level + superclass labels (HLE), and both (AHLE) labels.


We implement multiple variants of our model to analyze the impact of each semantic entity and the proposed regularization. \textbf{1) LME-MTL-S:} The multitask semantic embedding model learned with supercategories. \textbf{2) LME-MTL-A:} The multitask embedding model learned with attributes. \textbf{3) USE-No Reg.:} The unified semantic embedding model learned using both attributes and supercategories, without semantic regularization. \textbf{4) USE-Reg:} USE with the sparse coding regularization. 
%
We find the optimal parameters for the USE model by cross-validation on the validation set.
\begin{table}[h]
\footnotesize
\begin{center}
\resizebox{14cm}{!}{
\begin{tabular}{| m{1.2cm} | c || c c c | c c |}
\hline
&&\multicolumn{3}{c|}{Flat hit @ k (\%)} &\multicolumn{2}{c|}{Hierarchical precision @ k (\%)} \\
\hline
& Method & 1 & 2 & 5 & 2 & 5 \\
\hline
\multirow{2}{1.2cm}{ No semantics} & Ridge Regression & 38.39 $\pm$ 1.48 & 48.61 $\pm$ 1.29 & 62.12 $\pm$ 1.20 & 38.51 $\pm$ 0.61 & 41.73 $\pm$ 0.54\\
&LME & 44.76 $\pm$ 1.77 & 58.08 $\pm$ 2.05 & 75.11 $\pm$ 1.48 & 44.84 $\pm$ 0.98 & 49.87 $\pm$ 0.39 \\
\hline
\multirow{3}{1.2cm}{Implicit semantics} 
&ALE~\cite{ale} &  36.40 $\pm$ 1.03 & 50.43 $\pm$ 1.92 & 70.25 $\pm$ 1.97 & 42.52 $\pm$ 1.17 & 52.46 $\pm$ 0.37 \\
&HLE~\cite{ale} &  33.56 $\pm$ 1.64 & 45.93 $\pm$ 2.56 & 64.66 $\pm$ 1.77 & 46.11 $\pm$ 2.65 & \bf 56.79 $\pm$ 2.05  \\
&AHLE~\cite{ale} &  38.01 $\pm$ 1.69 &  52.07 $\pm$ 1.19 & 71.53 $\pm$ 1.41& 44.43 $\pm$ 0.66 & 54.39 $\pm$ 0.55 \\
\hline
\hline
\multirow{2}{1.2cm}{Explicit semantics} &LME-MTL-S & 45.03 $\pm$ 1.32 & 57.73 $\pm$ 1.75 & 74.43 $\pm$ 1.26 & 46.05 $\pm$ 0.89 & 51.08 $\pm$ 0.36 \\
&LME-MTL-A & 45.55 $\pm$ 1.71 & 58.60 $\pm$ 1.76 & 74.67 $\pm$ 0.93 & 44.23 $\pm$ 0.95 & 48.52 $\pm$ 0.29 \\
\hline
\multirow{2}{1.2cm}{USE}&USE-No Reg. & 45.93 $\pm$ 1.76 & 59.37 $\pm$ 1.32 & 74.97 $\pm$ 1.15 & 47.13 $\pm$ 0.62  & 51.04 $\pm$ 0.46 \\
&USE-Reg. & \bf 46.42 $\pm$ 1.33 & \bf 59.54 $\pm$ 0.73 & \bf 76.62 $\pm$ 1.45 & \bf 47.39 $\pm$ 0.82 & 53.35 $\pm$ 0.30 \\
\hline
\end{tabular}
}
\end{center}
\vspace{-0.2in}
\caption{\small Multiclass classification performance on \textbf{AWA-DeCAF} dataset (4096-D DeCAF features).}
\label{decafemulticlass}
\vspace{-0.2in}

\end{table}

\paragraph{Multiclass categorization.}
We first evaluate the USE framework for categorization performance. We report the average classification performance and standard error over 5 random training/test splits in Table~\ref{decafemulticlass}, using both flat hit@k, which is the accuracy at the top-k prediction made, and hierarchical precision@k from~\cite{devise}, which is a precision the given label is correct at $k$, at all levels.

The implicit semantic baselines, ALE-variants, underperformed even the ridge regression baseline with regard to the top-1 classification accuracy~\footnote{We did extensive parameter search for the ALE variants.}, while they improve upon the top-2 and hierarchical precision. 
This shows that hard-encoding structures in the label space do not necessarily improve the discrimination performance, while it helps to learn a more semantic space. 

Explicit embedding of semantic entities using our method improved both the top-1 accuracy and the hierarchical precision, with USE variants achieving the best performance in both. 
USE-Reg. made substantial improvements on flat hit and hierarchical precision @ 5, which shows the proposed regularization's effectiveness in learning a semantic space that also discriminates well.

\begin{table}[h]
\begin{center}
\resizebox{14cm}{!}{
\begin{tabular}{| p {1.2cm}  | m{8.5cm} | m{8.5cm} |}
\hline
Category & Ground-truth attributes & Supercategory + learned attributes \\ 
\hline
\hline
\centering
\includegraphics[width=1cm]{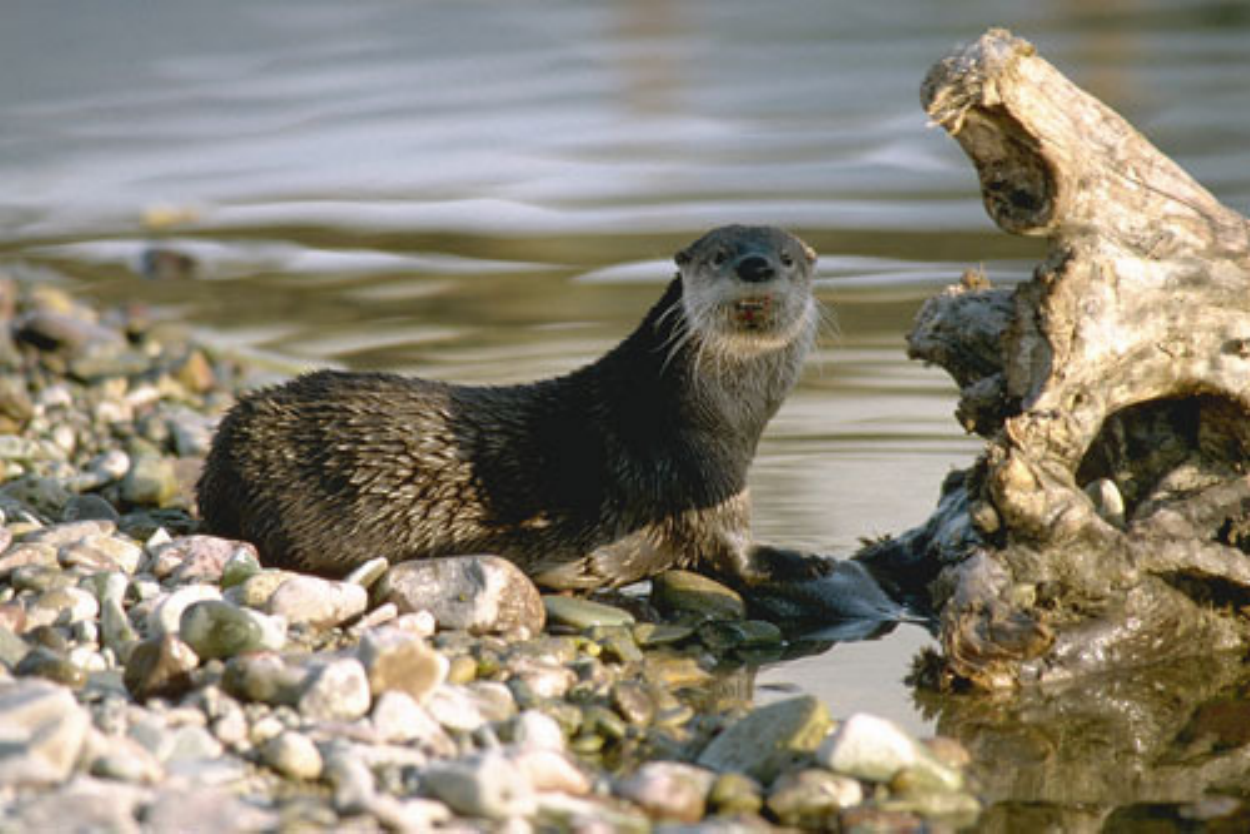} Otter & An animal that swims, fish, water, new world, small, flippers, furry, black, brown, tail, \dots&A musteline mammal that is quadrapedal, flippers, furry, ocean\\ 
\hline
\includegraphics[width=1cm]{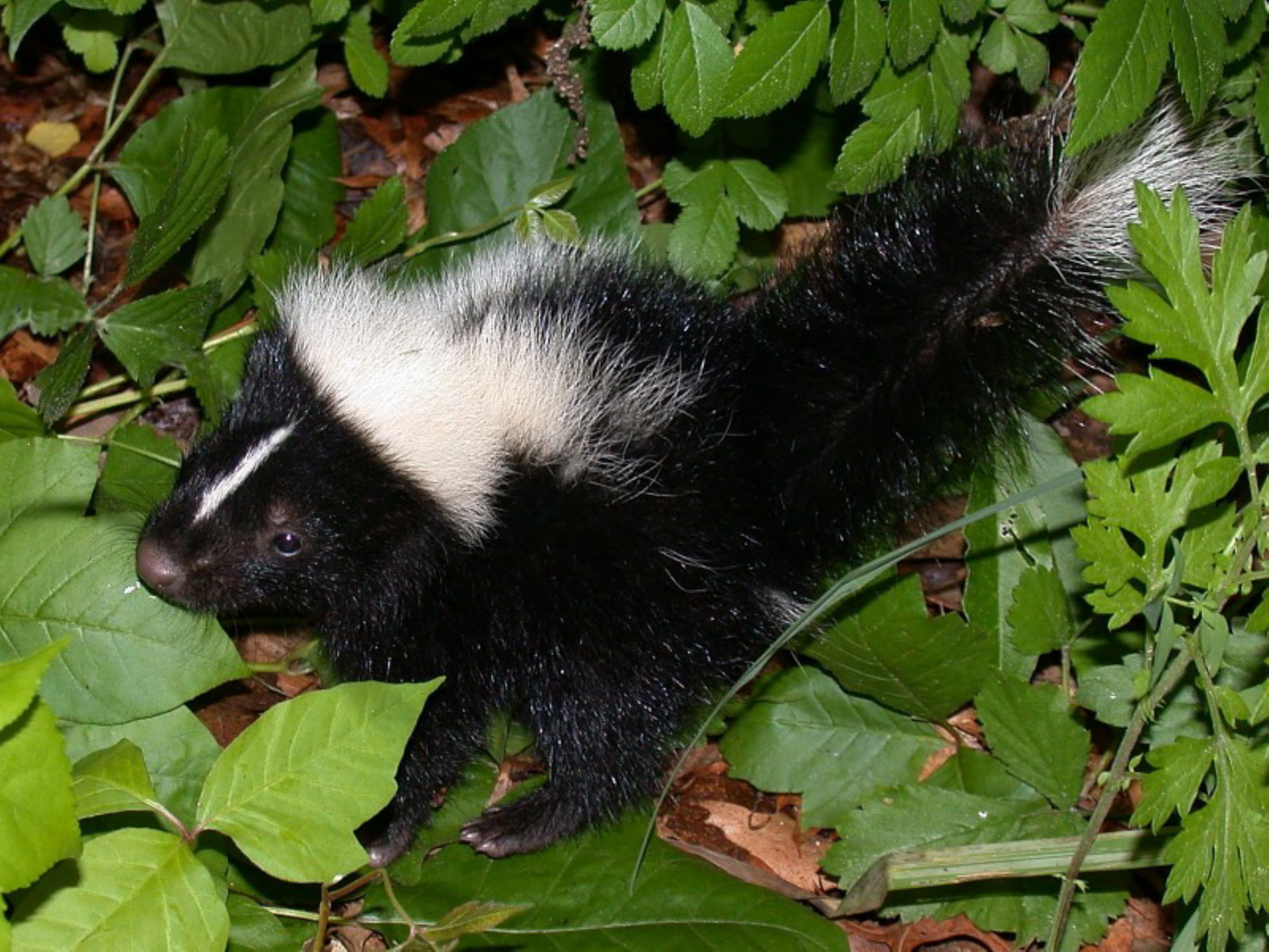} Skunk & An animal that is smelly, black, stripes, white, tail, furry, ground, quadrapedal, new world, walks, \dots & A musteline mammal that has stripes\\
\hline
\hline
\includegraphics[width=1cm]{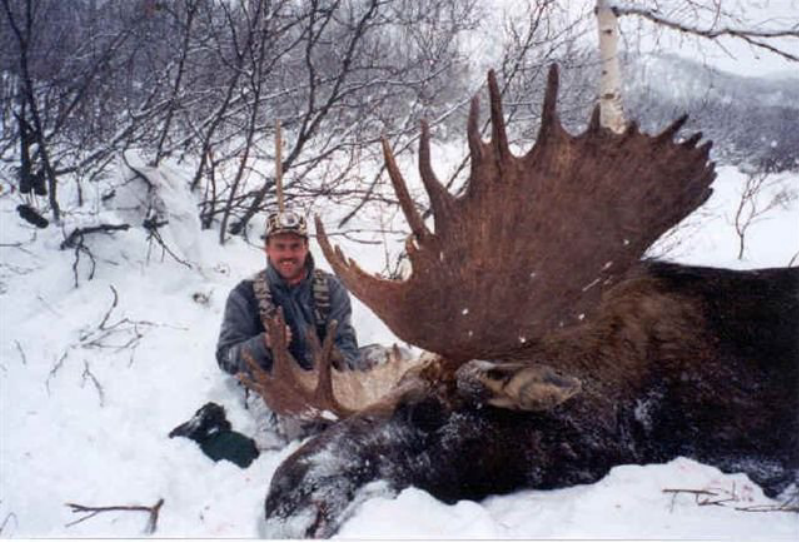} Moose & An animal that has horns, brown, big, quadrapedal, new world, vegetation, grazer, hooves, strong, ground,\dots &A deer that is arctic, \wrong{stripes}, black\\
\hline
\hline
Equine & N/A &An odd-toed ungulate, that is lean and active\\
\hline
\hline
Primate & N/A & An animal, that has hands and bipedal \\
\hline
\end{tabular}
}
\end{center}
\vspace{-0.15in}
\caption{\small Semantic description generated using ground truth attributes labels and learned semantic decomposition of each categorys. For ground truth labels, we show top-10 ranked by their human-ranked relevance. For our method, we rank the attributes by their learned weights. Incorrect attributes are colored in red.} \label{decomp}
\end{table}
\vspace{-0.15in}
\paragraph{Qualitative analysis.}
Besides learning a space that is both discriminative and generalizes well, our method's main advantage is its ability to generate compact, semantic description of each category it has learned. This is a great caveat, since in most models, including the state-of-the art deep convolutional networks, humans cannot understand what has been learned; by generating human-understandable explanation, our model can \emph{communicate} with the human, allowing understanding of rationale behind the categorization decision, and to possibly provide feedback for correction.

To show the effectiveness of using supercategory+attributes in the description, we report the learned reconstruction for our model, compared against the description generated by ground-truth attributes in Table~\ref{decomp}. The results show that our method generates compact description of each category, focusing on its \emph{discriminative} attributes. For example, our method selects \emph{flippers} for otter, and \emph{stripes} for skunk, instead of common nondescriminative attributes such as \emph{tail}. 
Further, our method selects attributes for each supercategory, while there is no provided attribute label for supercategories.

\vspace{-0.1in}
\paragraph{One-shot/Few-shot learning.}
Our method is expected to be especially useful for few-shot learning, by generating a richer description than existing methods that approximate the new input category using only trained categories, or attributes. For this experiment, we divide the $50$ categories into predefined $40/10$ training/test split. USE-Reg achieves the most improvement, improving two-shot result on AWA-DeCafe from 38.93\% to 49.87\%. Most learned reconstruction look reasonable, and fit to discriminative traits that help to discriminate between the test classes.

\bibliographystyle{unsrt}
\bibliography{strings,refs}
\end{document}